# Arduino Controlled Spy Robo Car using Wireless Camera With live Streaming

Nasir karim, Muhammad Ayoub Kamal, Mohsin Maknojia   , Saifan rafique

*Abstract*— Nowadays, the technological and digital world is developing very fast. Everything is getting smart, so we are talking about the technological world the devices like home appliances and other things are getting control by mobile applications, and this only happens by the device Arduino Uno / raspberry pi3 and many others. Still, in our research we have used Arduino Uno to create a Wi-Fi controlled car with camera-top on it to monitor everything in its surrounding, we have seen many similar projects which using Arduino to makes things easy to use and its saving time and energy too. Automation is used for operating an electronic device such as the remote control car, home lighting system, and other useful things or reduced human invention. This report proposes a design and implementation of a remote-controlled camera car by Wi-Fi technology mobile devices. In this analysis work, radio code and hardware technologies area unit used, like the wireless module of ESP8266 for (transmitter and receiver), Arduino Uno as microcontroller, associate H-bridge L293D IC for motor controller and 2 electrical DC motors are used to move the car, & a Camera attached on the top of the vehicle. (Abstract)

## I. INTRODUCTION

In the Modern world. There are many people who are investing in Wi-Fi technology. Wi-Fi technology used in such as airport, home, office, and other public areas. Wi-Fi technologies supported by the computer, laptop, game consoles, and smartphones. Wi-Fi technology helps people to work and communicate without network cabling. This is very helpful for many users. Nowadays, Wi-Fi technology is not for internet use. Wi-Fi technology can control any kind of equipment like Aircondition, Television, Alarm, and many other appliances that support Wi-Fi technology. Hence it is possible to control the robot with Wi-Fi technology. This project is to design a robotic vehicle that can be utilized in defense. This robot is controlled by smartphone and hardware. The remote camera is on a vehicle for checking the circumstance around the robot. This kind of robot is helpful in any spying reason field like military, and the police further it can be used for security of assets. Another benefit of this robot of this size can ignore people. It has endless applications and can be utilized as a part of various situations and environments. The use of this robot that can command is sent to a robot for controlling the operation of the robot to moves left, right, forward, or in reverse. The receiving command from application to the Wi-Fi module is connected with the Arduino to receiving control and motor with an Arduino via motor driver where they can help for movement of the vehicle. The power of the DC source is taken from batteries attached to a robot for Arduino, WI-Fi, camera, and motor driver.

## II. RELATED WORK

So far, many researchers have been done and published in the field of Arduino based projects for different parameters used in many different sectors. Automation is to control the systems and technologies to save time and energy. This project comprises an Arduino Uno with the Seed Studio GPRS Shield V2.0 related to an ultrasonic sensor HC-SR04. The thought is behind this is particularly direct; when the ultrasonic sensor distinguishes a qualification out there that are evaluating, I will get a methodology my supportive. I would then be able to put a get back to "alarm" the framework.[1] In a previous project, we discuss the security system alarm by using sensors. But now to add some advancement to the project, now we are using the camera to capture the person image; in previous, we can only get alert by someone is coming through but in this, the device can achieve the vision by which we can know who is behind the mess. So we are going to utilize Arduino Bluetooth Camera with the help of ultrasonic to detect whether a stranger has entered our home and catch a photograph of him consequently once he gets into the region of the ultrasonic wave[2]. There is another similar project of RC robocar in this. The function of this robot car is easy to identify by its name. Truly! You can capture pictures of the encompassing condition by the camera, and transfer video to your Android application. In the meantime, you can control it with your android application. Advantageous and just. We include a dish tilt for giving more extensive view scope of the car. It moves quickly and responds rapidly. You can change the course of care, as indicated by the video. So in the event that you need to "investigate" some spot slender and may harm to your wellbeing, you would remote be able to control your car as motion picture appeared. Obviously, it for playing as well as examining[3] Control from the iPhone ios application by means of Bluetooth connection. Control the GoPro camera via Wi-Fi and suspension using a servo motor. Straightforward FPV mode. Likewise, have Obstacle shirking autopilot with ultrasonic sensor and gyro.[4] This is a simple four-wheeled car that is controllable via Bluetooth connection. The vehicle can hold any mobile phone, which can be used as

Muhammad Ayoub Kamal is with MIIT University, Kaula lumpur Malaysia (Email: engr.muhammadayoubkamal@gmail.com)
Nasir karim is with Cloud Computing Software House, Hyderabad, Pakistan (Email: Nasirmomin95@gmail.com)
Mohsin Maknojia is with Cloud Computing Software House, Hyderabad, Pakistan (Email: Mohsinmaknojia19953@gmail.com)
Saifan rafique is with with Cloud Computing Software House, Hyderabad, Pakistan. (Email: Maredia_saifan@gmail.com)

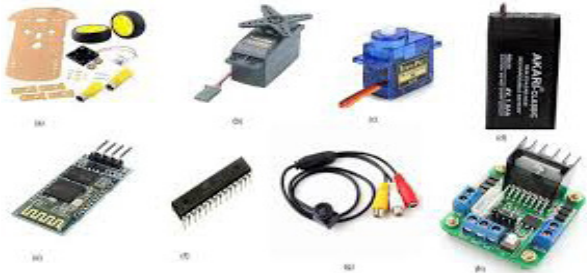

Fig. 1. Basic Idea of Project

an IP camera. So we decided to make this robot with JavaScript using a node webserver where we can monitor where the robot goes [5]. And then, this technology is also used in an ultrasonic sensor The initial step is to install the DC motor and two wheels on the baseboard, and the link in DC motor is connected operating at a profit line underneath the red line. Install the Arduino motherboard and install it cautiously as per the position saved by the motherboard. Install the battery holder. Connect the connector to the Arduino sensor shield v5 control interface. The Drive Board Module is installed on the baseboard, and the Drive Board Module is a crucial step that controls the task of two motor capacities. Connect the Arduino sensor shield v5 with the Drive Board Module. Install the ultrasonic sensor on the Drive Board Module. Connect the connector to the right port in the request of the interface of the schematic diagram[6].

### III. MATERIAL AND METHOD

The main component of the spy camera car system is the WIFI module X1. WIFI module X1 like a brain for RC camera cars. I used a WIFI module X1 to establish a connection between the car and my application. The camera is connected with PanTilt X1 when the user gives rotation command from the application the WIFI module X1 instruction and sends the instruction to Pan-Tilt X1 to rotate the camera and WIFI module X1 transferring the video stream to the application.

REQUIREMENT COMPONENTS :

- ARDUINO UNO
- MOTOR BASED SHIELD
- WIFI MODULE
- USB CAMERA
- VEHICLE CHASSIS
- PAN TILT
- BATTERY

The important part of the RC car is chassis, fix the holder on the chassis, and now interface red link to the VIN of Crowtial-motor base shield, the dark link associate with the GND of the motor base shield. Now this time to power up and test the motor are working or not. At that point, interface him

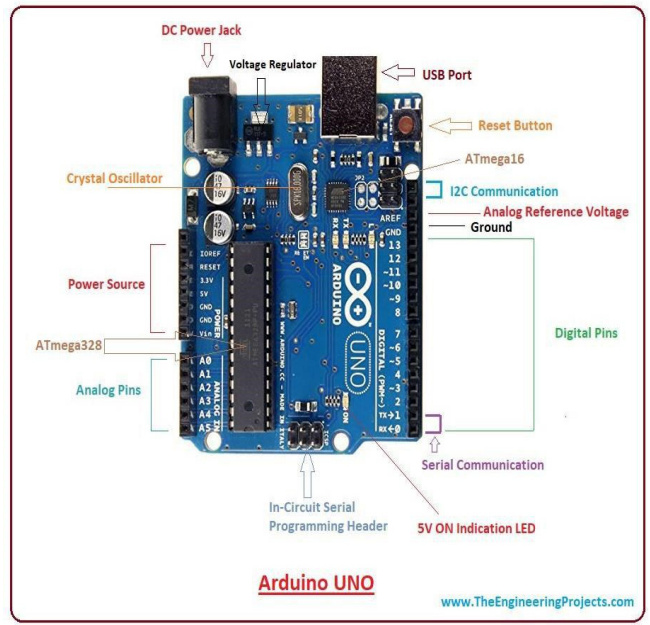

Fig. 2. Arduino Uno

CrowduinoUNO with USB connection to PC and Upload the test code. On the off chance that the vehicle wheel can't keep running same way trade the two-wire of engine if the Motor 1 go ahead, yet the Motor 2 return, Now we simply change the link of engine 2, the red link to OUT2, the dark link to OUT1, at that point test once more. Every wheel is running well, so now we need the camera that can move in mufti direction with the help of pan-Tilt combine with the pan-Tilt with the camera. We install Pan-Tilt with the chassis, make four holes and fix. The first thing we want to Plug the camera onto the Pan-Tilt and interface the WLAN module to the UART port of Crowtial-Motor base defend. Presently the Connect the Pan-Tilt to the Crowtial-Motor base defend, up-servo connect with D7 and down-servo connect with D6. Then connect the camera to the WLAN module with a USB link and transfer the code. We need to the application that can run the RC car, so install the app on smartphone and power the RC car, then open Wi-Fi setting phone search for Wi-Fi signal "Electro" connect it without any password [7-10].

A - ARDUINO UNO
The ASCII text file Arduino atmosphere permits the user to put in writing code and transfer it to the I/O board. The environment is written in Java. Arduino development atmosphere contains a text editor for writing code, message space, text console, and toolbar with buttons for common functions, and a series of menus. It connects to the Arduino hardware to transfer programs and communicate with them. Arduino programs are written in C or C++. Arduino options, capable of compilation and uploading programs to the Board with one click. Software written using Arduino are called sketches.

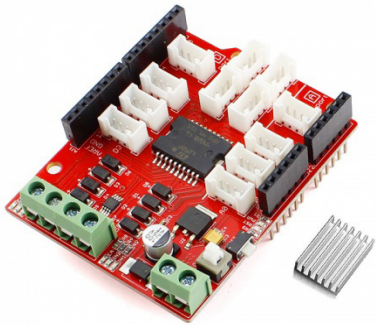

Fig.3.  Motor based shield

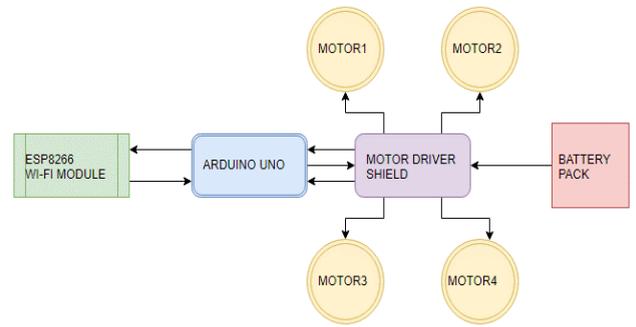

Fig. 4.  System block diagram

### B - MOTOR BASED SHIELD

The motor base shield is based on the dual-channel motor driver IC L298P, and it incorporates the Crowtail-base shield and motor and stepper shield. It will carry accommodation for your work with Crowduino. Stackable plan as yet utilizing, The change controls on/off the Motor Base Shield, which make it progressively helpful in utilizing, particularly in your building& troubleshooting. To guarantee the L298P works in the best status, the PCB format meets the maximum current 2A necessities. There are 8 LEDs to show the Power, RST, Speed, Directions, to enable the client to get the point by point data of working status.

### C – WIFI MODULE

At first, Wi-Fi was utilized as opposed to just the 2.4GHz 802.11b standard, anyway, the Wi-Fi Coalition has expanded the standard use of the Wi-Fi term to incorporate any sensible framework or WLAN thing subject to any of the 802.11 pointers, just as 802.11b, 802.11a, two-fold band and so on, endeavoring to avert disarray about remote neighborhood capacity. Agonizing about its typical operating zone of within 100 meters, it's eminently helpful in-home and works environment conditionsconditions. Express gratefulness for Wi-Fi development and different near strategies, with exciting addition in Smartphone customers, smartphones have bit by bit improved into a generally helpful advantageous contraption, what is more, offered individuals to their step by step use. Starting late, partner degree ASCII content file organize mechanical man has been generally utilized in smartphones. Android has all out programming bundle pack containing an operating system, middleware layer, and focus applications. Particular in significance different existing stages like iOS (iPhone OS), it goes with programming bundle improvement pack (SDK), which gives essential instruments and Application. Using a Smartphone on the grounds that the "mind" of a system is starting at as of now a dynamic examination field with various open probabilities and promising potential results during this task we will in general present a study of current robots stressed by cell phone and analyze a shut circle control structure.

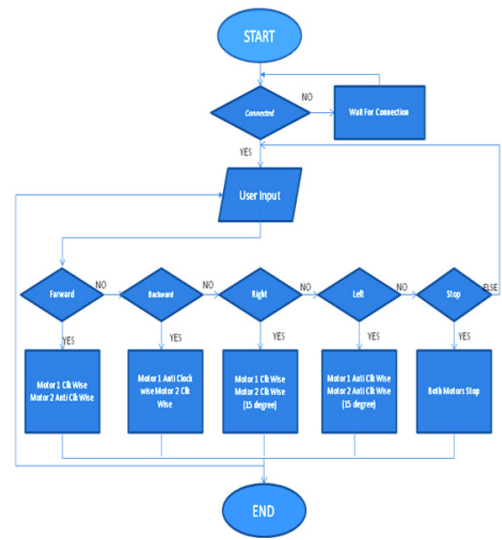

Fig. 5.  System Dataflow  diagram

### IV. SYSTEM BLOCK DIAGRAM

Give us a chance to take a Wifi module squirming electron in one area. This squirming will electron cause a wave impact, to some degree the same as dropping a rock in a lake. The impact is an electromagnetic (EM) wave, which goes out from the underlying area of those outcomes in electrons to squirm in remote areas. An RF collector can identify the distant electron squirming. The RF correspondence framework, at that point, further uses this wonder by squirming electrons in a particular example in order to speak to data. The recipient can make similar data accessible at a remote area by building up a correspondence without any wires. In a large portion of the remote frameworks, an architect has two abrogating requirements: it must work over a specific separation (range) and move a specific measure of data inside a time period (information rate).falsely made radio waves that sway at different picked frequencies. RF correspondence is utilized in numerous ventures, including TV broadcasting, radar frameworks, PC and versatile stage systems, remote control, remote metering/checking, and numerous more. While individual radio parts, for example, blenders, channels, and power speakers can be characterized as indicated by working recurrence run, they can't be carefully sorted by remote standard (e.g.Wi - Fi, Bluetooth, and so forth.) on the grounds that these gadgets just give physical layer (PHY) support . conversely, RF modules, handsets, thus Cs regularly incorporate information connection layer support for at least one remote correspondence conventions

## V. RESULTS

It tends to be constructed further to function as a HUMANOID. It can have numerous utilizations in handy fields from a young person's robots to robots working in enterprises. It is useful in wars as a piece of spying. The proposed robot can be additionally improved as far as choice taking capacities by utilizing fluctuated sorts of sensors and along these lines could be utilized in huge ventures for various applications.

The future enhancement of remote control car with the camera is the following:

- Remote control car is used in bomb-squads to defuse or detonated explosives.
- Used in movie making.
- Used in the radiological survey.

## VI. CONCLUSION

In this Research paper, we have a spy robocar that is used for the different field for surveillance. The paper contains information about the overall applications of the device to monitor and secure places that are difficult to reach, for instance, in the case of an investigation by intelligent agencies. It aims to provide accuracy and constant monitoring of the vacancy from any threat or break-ins via an embedded system and camera that sends information to the other end of the observer entity.

REFERENCES

[1] Montrose, M. I. (1996). Printed circuit board design techniques for EMC compliance (Vol. 1, p. 996). Piscataway, NJ: IEEE Press.
[2] Forouzan, B. A., & Fegan, S. C. (2006). TCP/IP protocol suite (Vol. 2). McGraw-Hill.
[3] Balakrishnan, H., Stemm, M., Seshan, S., & Katz, R. H. (1997). Analyzing stability in wide-area network performance. ACM SIGMETRICS Performance Evaluation Review, 25(1), 2-12.
[4] Briscoe, N. (2000). Understanding the OSI 7-layer model. PC Network Advisor, 120(2).
[5] G.Huston, Analyzing the Internet BGP routing table, Internet Protocol Journal 4 (1) (2001).
[6] McKeown, N. (2009). Software-defined networking. INFOCOM keynote talk, 17(2), 30-32.
[7] Swales, A. G., Papadopoulos, A. D., & Tanzman, A. (2001). U.S. Patent No. 6,233,626. Washington, DC: U.S. Patent and Trademark Office.
[8] Voellmy, A., & Wang, J. (2012). Scalable software-defined network controllers. ACM SIGCOMM Computer Communication Review, 42(4), 289-290.
[9] International Journal of Applied Information Systems (IJAIS) – ISSN: 2249-0868 Foundation of Computer Science FCS, New York, USA Volume 11 – No. 7, December 2016 – www.ijais.org 10 Introduction to Software Defined Networks (SDN)
[10] Indian Journal of Science and Technology, Vol 10(29), DOI: 10.17485/ijst/2017/v10i29/112447, August 2017